\documentclass[11pt]{article}

\usepackage[preprint]{acl}

\usepackage{times}
\usepackage{latexsym}

\usepackage[T1]{fontenc}

\usepackage[utf8]{inputenc}
\usepackage{amsmath}
\usepackage{amssymb}
\usepackage{hyperref}

\usepackage{enumitem}

\usepackage{microtype}

\usepackage{inconsolata}

\usepackage{pifont}
\usepackage{xcolor}
\usepackage{array}

\usepackage{url}

\newcommand{\cmark}{\textcolor{green!60!black}{\ding{51}}}%
\newcommand{\xmark}{\textcolor{red!75!black}{\ding{55}}}%
\newcommand{\pmark}{\textcolor{orange!85!black}{\(\triangle\)}}%

\usepackage{graphicx}

\usepackage{booktabs}
\usepackage{multirow}
\usepackage{xcolor}
\usepackage[table]{xcolor}
\usepackage[most]{tcolorbox}

\usepackage[most]{tcolorbox}

\newtcolorbox{PlannerPromptBox}{
  breakable,
  colback=white,
  colframe=blue!30,
  fonttitle=\bfseries,
  title=Planning Module Prompt,
}

\newtcolorbox{PromptforBrickPoseEstimation}{
  breakable,
  colback=white,
  colframe=orange!35,
  fonttitle=\bfseries,
  title=Brick Pose Estimation Trajectories,
}

\newtcolorbox{PromptforBrickSelection}{
  breakable,
  colback=white,
  colframe=teal!30,
  fonttitle=\bfseries,
  title= Brick Selection Trajectories,
}

\newtcolorbox{PromptforLearningwithWorldFeedback}{
  breakable,
  colback=white,
  colframe=blue!30,
  fonttitle=\bfseries,
  title=Learning with Feedback Trajectories,
}

\newtcolorbox{InitInteractPromptBox}{
  breakable,
  colback=white,
  colframe=blue!30,
  fonttitle=\bfseries,
  title=Planning Prompt: Initialize Shallow UI Interaction,
}

\newtcolorbox{StateClassifyPromptBox}{
  breakable,
  colback=white,
  colframe=orange!35,
  fonttitle=\bfseries,
  title=Feedback Prompt: State Classification,
}

\newtcolorbox{TerminalPromptBox}{
  breakable,
  colback=white,
  colframe=orange!35,
  fonttitle=\bfseries,
  title=Feedback Prompt: Terminal State Verification,
}

\newtcolorbox{PlanErrorPromptBox}{
  breakable,
  colback=white,
  colframe=orange!35,
  fonttitle=\bfseries,
  title=Feedback Prompt: Error Attribution (Planning),
}

\newtcolorbox{ActionErrorPromptBox}{
  breakable,
  colback=white,
  colframe=orange!35,
  fonttitle=\bfseries,
  title=Feedback Prompt: Error Attribution (Action),
}

\NewDocumentCommand{\heng}
{ mO{} }{\textcolor{red}{\textsuperscript{\textit{Heng}}\textsf{\textbf{\small[#1]}}}}

\NewDocumentCommand{\cheng}
{ mO{} }{\textcolor{orange}{\textsuperscript{\textit{cheng}}\textsf{\textbf{\small[#1]}}}}

\title{Brick-Composer: MLLMs Construct Everything from Building Blocks }

\newcommand{\logoh}{1.35em}

\author{\normalsize Jiateng Liu\textsuperscript{1}, Bingxuan Li\textsuperscript{1}, Zhenhailong Wang\textsuperscript{1}, Rushi Wang\textsuperscript{1}, Kaiwen Hong\textsuperscript{1}, Cheng Qian\textsuperscript{1},\\ \normalsize \textbf{Jiayu Liu\textsuperscript{1}, Denghui Zhang\textsuperscript{2}, Katherine Driggs-Campbell\textsuperscript{1}, Manling Li\textsuperscript{3}, Heng Ji\textsuperscript{1}}
\\[0.4em] \normalsize UIUC\textsuperscript{1}, Stevens Institute of Technology\textsuperscript{2}, Northwestern University\textsuperscript{3}
 \\[0.25em] \normalsize \raisebox{-0.2\height}{\includegraphics[height=\logoh]{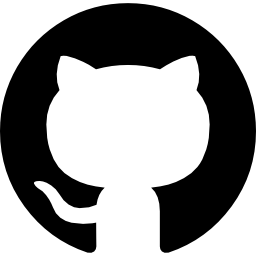}} Project Page: \url{https://github.com/Lumos-Jiateng/Brick-Composer}  }

\begin{document}
\maketitle

\begin{abstract}

We dream of AI agents that can read arbitrary designs and construct real-world objects from reusable building blocks. As a first step toward this vision, we study whether multimodal large language models (MLLMs) possess the visual grounding and spatial reasoning capabilities required for brick assembly. We formulate brick assembly as a sequential decision-making problem, where each step involves two subtasks: \textit{brick selection}, identifying the target brick from candidate components, and \textit{brick pose estimation}, predicting where and how the selected brick should be placed. To support this study, we introduce \textbf{BC-Bench} (\textbf{B}rick \textbf{C}onstruction \textbf{Bench}mark), the first benchmark for evaluating MLLMs on assembly with diverse bricks. Experiments show that current state-of-the-art MLLMs remain far from reliable builders, struggling with fine-grained brick selection and failing at precise pose estimation. To bridge this gap, we propose \textbf{Brick-Composer}, a learning framework that equips MLLMs with assembly skills through three complementary signals: \textit{Human Design Sparks}, which provide affordance-rich construction demonstrations; \textit{World Feedback}, which grounds predicted actions in visual and physical consequences; and \textit{Synthetic Experience}, which scales learning beyond existing object designs. Brick-Composer improves brick selection accuracy by over three times, substantially reduces pose estimation errors, and raises strict step-level assembly success from less than 1\% to around 15\%. After training, a Qwen-3-8B can correctly compose up to 42\% of the steps for a complete object, suggesting that MLLMs can acquire assembly capabilities through targeted, physically grounded learning.



\end{abstract}



\section{Introduction}

\begin{figure*}[t]
    \centering
    \includegraphics[width=0.95\textwidth]{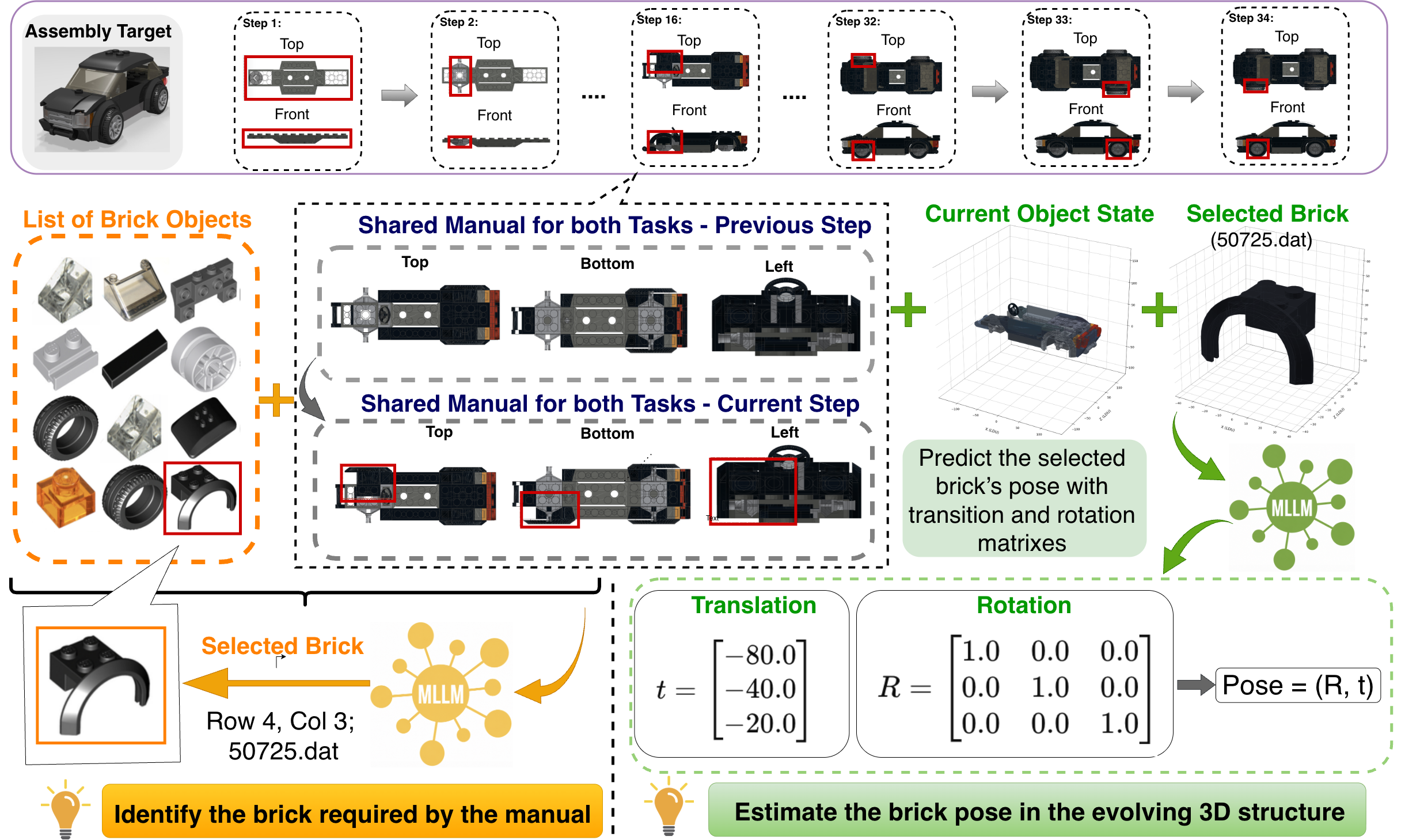}
    \caption{
Overview of the BC-Bench task setting.
\textbf{Left: Brick selection.} The model selects the required brick from a candidate grid using manual images.
\textbf{Right: Brick pose estimation.} Given the manual context, current assembly state, and selected brick, the model predicts the brick's target pose as a translation vector and rotation matrix.}
    \label{fig:task}
\end{figure*}

It is such a beautiful dream to imagine a future where robots can read arbitrary designs and simply construct almost anything from reusable building blocks: a piece of furniture, a car, a house, or even an aircraft. Yet realizing this dream, and achieving such a level of generalization, requires a surprisingly deep form of intelligence that goes far beyond visual understanding alone, and we are still only at the very early stages of reaching it~\cite{pun2025brickgpt,tie2025manual2skill,kulits2026bricknet,jing2026assemlm}. A capable assembly agent must recognize building blocks from fine-grained geometric features and functional affordances, infer each brick’s target pose within the correct assembly sequence, execute reliable actions in the physical world, and continuously monitor the long-horizon construction process while remaining able to recover from errors along the way.

In this paper, we use LEGO-style assembly as a proxy task for pursuing this long-term vision, taking an initial step toward enabling multimodal large language models (MLLMs) to reason about complex object construction. As illustrated in Figure~\ref{fig:task}, we formulate assembly with diverse bricks as a sequence of step-wise decisions. At each step, the model observes the instructional manual and is required to solve two coupled subtasks: brick selection, which identifies the next brick from a set of candidates, and brick pose estimation, which predicts where and how the selected brick should be placed based on the current assembly state and the brick's geometry. This setting closely reflects how humans follow assembly manuals: interpreting visual instructions, choosing the appropriate component, and integrating it into an evolving structure with the correct spatial pose.

To support this study, we introduce \textbf{BC-Bench} (\textbf{B}rick \textbf{C}onstruction \textbf{Bench}mark), the first benchmark for evaluating MLLMs on assembly with diverse bricks. Built from human-designed LEGO-style objects and diverse brick inventories, BC-Bench uses a custom simulator to reconstruct step-wise assembly processes and render multimodal observations, including manual-style instruction views, candidate brick views, and evolving assembly states from orthogonal and isometric viewpoints.\footnote{We respect designers' rights, follow responsible content-use practices, and avoid large-scale web scraping.} This design enables controlled evaluation of brick selection and pose estimation while preserving the structural complexity, design diversity, and real-world affordances of human-created assemblies. Using BC-Bench, we conduct a comprehensive evaluation of whether current MLLMs can perform LEGO-style assembly with diverse bricks, covering both fine-grained brick selection and brick pose estimation. The results reveal a clear gap between current model capabilities and the demands of reliable assembly: even state-of-the-art MLLMs struggle to identify the correct brick at each step, and most models produce large translation and rotation errors when asked to place the selected brick.

Despite these limitations, we introduce \textbf{BrickComposer}, a learning framework that equips MLLMs with assembly skills through three complementary forms of supervision: \textit{Human Design Sparks}, \textit{World Feedback}, and \textit{Synthetic Experience}. Together, these signals capture three key merits of assembly learning: design-driven structures grounded in real-world affordances, simulator-based feedback for error discovery and recovery, and scalable compositional diversity beyond existing object designs. \textit{Human Design Sparks} provide high-quality demonstrations from human-designed objects, showing how diverse bricks are used to form meaningful structures and support different object functions. While these examples teach the basic language of assembly, they are valuable but scarce: 80 human-designed objects yield only about 3K step-wise training examples. \textit{World Feedback} grounds model predictions in their visual and physical consequences by executing predicted selections or poses in our simulator, rendering the resulting assembly state, and highlighting the predicted brick or placement. This exposes model errors and supports recovery, either through inference-time revision or training-time correction. \textit{Synthetic Experience} scales learning beyond existing designs by generating physically plausible brick configurations through feasible attachments, while filtering for collision-free and compact structures. We produce synthetic objects with 20 to 100 pieces and yields around 40K step-wise training examples, exposing models to broader geometries, connection affordances, and attachment patterns.

Together, these components provide a progressive path for assembly learning: from interpreting human designs, to correcting simulated actions, and generalizing across scalable physically grounded configurations. Empirically, Brick-Composer turns LEGO-style assembly from an almost impossible zero-shot challenge into an emerging capability. Step-wise brick selection accuracy increases from roughly 23\% with base Qwen-3 models to around 70\% after learning, while average translation and rotation errors are significantly reduced . Under a strict criterion requiring both the correct brick and an accurate pose, base MLLMs achieve nearly 0\% successful assembly steps, whereas our learned models can reach an average of 15\% and peaks at 42\%. This indicates that MLLMs can already complete a non-trivial subset of construction steps by selecting the right component and placing it into an acceptable target pose. Although fully autonomous assembly remains far from solved, these results suggest that assembly is a learnable capability: with larger-scale human supervision, stronger visual feedback, and higher-quality physically grounded data, MLLMs may gradually move from reading assembly instructions to executing them reliably.


\section{Related Work}


\paragraph{Spatial intelligence of MLLMs.}
Recent work has studied spatial intelligence as a key capability of MLLMs. 3D-LLM~\citep{hong2023_3dllm} and SpatialVLMs~\citep{chen2024spatialvlm} show that VLMs remain limited in spatial reasoning and improve this ability by injecting 3D scene data and large-scale spatial visual question answering data. Subsequent studies extend this direction to grounded region-level reasoning~\citep{cheng2024spatialrgpt}, depth-aware spatial understanding~\citep{cai2024spatialbot}, metric estimation and 3D grounding~\citep{daxberger2025mmspatial}, video-based spatial intelligence~\citep{yang2025thinking}, embodied-agent evaluation~\citep{yang2025embodiedbench}, and fine-grained spatial cognition benchmarks~\citep{xu2025spatialbench}. These studies demonstrate the growing attention for spatial reasoning ability of MLLMs, motivating our work on assembly-specific reasoning with diverse bricks, where a model must identify the target brick and estimate its placement pose within an evolving structure.

\paragraph{MLLMs for brick assembly and manufacturing.}
MLLMs for assisted brick assembly and manufacturing have been studied only recently. BrickGPT~\citep{pun2025brickgpt} generates physically stable brick structures from text by fine-tuning a LLM, but it relies on simplified brick primitives, leaving open how models handle diverse parts with varied geometry and affordances. Recent LEGO-oriented work, including BrickNet~\citep{kulits2026bricknet} and LEGO Co-builder~\citep{huang2025legocobuilder}, expands the brick vocabulary and studies build-sequence generation or assembly-state analysis, but does not directly target visually grounded assembly assistance. AssemLM~\citep{jing2026assemlm} further studies spatial MLLMs for robotic assembly, highlighting the importance of 3D geometric reasoning in assembly tasks. Our work is the first to address the lack of systematic evaluation of MLLMs for fine-grained brick selection and single-step pose estimation in realistic assembly settings by introducing a diverse benchmark and scalable learning pipeline for visually grounded brick assembly reasoning.

\paragraph{Robotics for brick assembly.}
Robotic assembly has long been studied as a challenging problem. Classical assembly settings require precise pose alignment and often rely on compliance control or contact feedback, while recent learning-based systems study assembly through manual understanding, 6D pose estimation, and robot policy learning. For example, Manual2Skill~\citep{tie2025manual2skill} and Manual2Skill++~\citep{tie2025manual2skillpp} learn executable skills from manuals; robotic LEGO assembly has been studied from human demonstrations~\citep{liu2023roboticlego}; pose-centric methods estimate 6D object or template poses from RGB-D or point-cloud observations~\citep{deng2020self,stevsic2020learning}; and recent VLA-style methods directly predict robot-specific grasp actions or action policies~\citep{mahler2017learning,zhen2024threedvla,singh2025ogvla,kim2025openvla}. Pose estimation and action execution remain core bottlenecks in robotic assembly. In contrast, our work opens a complementary angle for using MLLMs to select bricks and infer their target poses, providing an interpretable and simulation-evaluable step toward downstream robotic execution.

\section{Benchmarking MLLMs for Assembly}
\label{sec:settings}



\begin{table*}

\centering
\caption{
Comparison of BC-Bench with related assembly and LEGO-style benchmarks.
\cmark indicates the feature is explicitly supported; \xmark indicates it is not the focus;
\pmark indicates partial or related support.
}
\label{tab:benchmark_comparison}
\setlength{\tabcolsep}{4.2pt}
\renewcommand{\arraystretch}{1.12}
\resizebox{\linewidth}{!}{
\begin{tabular}{lcccc}
\toprule
\textbf{Benchmark / Dataset} 
& \textbf{Domain} 
& \textbf{Diverse Parts} 
& \textbf{Manual-based} 
& \textbf{Pose Est. w/ Rotation} \\
\midrule

LEGO Manual~\citep{wang2022visuallego}
& LEGO Manual Understanding
& \pmark
& \cmark
& \xmark \\

LEGO-Puzzles~\citep{tang2025legopuzzles}
& LEGO Spatial Reasoning
& \xmark
& \pmark
& \xmark \\

BrickGPT~\citep{pun2025brickgpt}
& Text-to-brick Generation
& \xmark
& \xmark
& \xmark \\

LEGO Co-builder~\citep{huang2025legocobuilder}
& LEGO assembly Assistant
& \pmark
& \cmark
& \xmark \\

BrickNet~\citep{kulits2026bricknet}
& Brick Sequence Generation
& \cmark
& \xmark
& \xmark \\

AssemLM~\citep{jing2026assemlm}
& Furniture Assembly
& \pmark
& \cmark
& \pmark \\

Manual2Skill~\citep{tie2025manual2skill}
& Furniture Assembly
& \xmark
& \cmark
& \pmark \\

\rowcolor{gray!12}
\textbf{BC-Bench (Ours)}
& Diverse LEGO Brick Assembly
& \cmark
& \cmark
& \cmark \\
\bottomrule
\end{tabular}
}
\end{table*}

We consider a step-wise brick assembly scenario in which an MLLM serves as an assembly agent in a simulated construction environment. Given a complex target object, the model observes the corresponding assembly manual, the current partially assembled structure, and a workspace containing candidate bricks with diverse geometries, functions, and physical affordances. As shown in Figure~\ref{fig:task}, each assembly step requires the model to solve two coupled problems. First, in \textbf{\textit{Brick Selection}}, the model identifies the brick specified by the current manual instruction from the available candidate set. Second, in \textbf{\textit{Brick Pose Estimation}}, the model predicts the selected brick's 3D placement, including its orientation and position relative to the existing structure. The process repeats iteratively until the target object is completed. We compare our benchmark with existing assembly, LEGO, and spatial reasoning benchmarks in Table~\ref{tab:benchmark_comparison}. We give more benchmark specific details in Appendix~\ref{app:dataset}.

\subsection{Brick Selection}

As shown in Figure~\ref{fig:task} (Left), the brick selection task evaluates whether an MLLM can understand the visual assembly manual and identify the required brick from its geometric features and functional affordances. At each step, the model is given the assembly manual, including current-step and target-step views rendered from six orthogonal perspectives, together with a visual grid of candidate bricks. We formulate brick selection as grounded localization over this grid, where the model predicts the row and column of the target brick. Formally, at assembly step $t$, let $M_t$ denote the manual context, including the current-step view and the target-step view. Let the candidate brick grid be denoted as
\begin{equation}
    \mathcal{B}_t = \{b_t^{i,j} \mid 1 \leq i \leq H, 1 \leq j \leq W\},
\end{equation}
where $b_t^{i,j}$ represents the candidate brick located at row $i$ and column $j$ in a grid with $H$ rows and $W$ columns. The ground-truth target brick is represented by its grid coordinate:
    $y_t^\ast = (i_t^\ast, j_t^\ast).$
Given the manual context $M_t$, the candidate grid $\mathcal{B}_t$, and a task prompt $P_{\mathrm{1}}$, the MLLM parameterized by $\theta$ predicts the grid coordinate of the brick required for the current step:
\begin{equation}
    \hat{y}_t = (\hat{i}_t, \hat{j}_t)
    =
    \mathrm{MLLM}_{\theta}(P_{\mathrm{1}}, M_t, \mathcal{B}_t).
\end{equation}
The selected brick is therefore given by $\hat{b}_t = b_t^{\hat{i}_t,\hat{j}_t}$. This task captures a common requirement in assembly. The challenge of the task increases when the workspace contains many visually similar parts that share colors or shapes but differ in subtle geometry or affordances, which are analyzed in Section~\ref{sec:experiments}.

\subsection{Brick Pose Estimation}

As shown in Figure~\ref{fig:task} (Right), the brick pose estimation task evaluates whether an MLLM can infer how the selected brick should be placed in the current state under the correct environmental scale. At each step, the model is given the manual, the partially assembled structure, and the geometric details of the selected brick. The model must predict the brick's translation and rotation in the target-object coordinate system. Formally, at assembly step $t$, let $M_t$ denote the manual context, including the current-step view and the target-step view. Let $S_t$ denote the current partially assembled structure in the simulated environment, represented by multi-view renderings under the target-object coordinate system $\mathcal{C}$. As shown in Figure~\ref{fig:task}, this includes six orthogonal views and one isometric view:
$S_t=\{v_t^{1}, v_t^{2}, \ldots, v_t^{K}\},$ where each $v_t^{k}$ is a 2D rendering from a predefined camera view, and $K=7$ in our setting. This representation exposes the scale and spatial layout of the current structure. Similarly, let $b_t$ denote the selected brick, whose geometry is rendered in its intrinsic coordinate system $\mathcal{C}_{b_t}$ using the same multi-view format: $b_t=\{g_t^{1}, g_t^{2}, \ldots, g_t^{K}\}$, where each $g_t^{k}$ shows the selected brick from the corresponding view, revealing its geometry, connection surfaces, and orientation-sensitive affordances. The goal is to predict the absolute pose of the selected brick in $\mathcal{C}$. We represent the ground-truth pose as a rigid transformation:
\begin{equation}
    P_t^\ast =
    \begin{bmatrix}
        R_t^\ast & T_t^\ast \\
        \mathbf{0}^{\top} & 1
    \end{bmatrix}
    \in SE(3),
\end{equation}
where $R_t^\ast \in SO(3)$ is the ground-truth rotation matrix and $T_t^\ast \in \mathbb{R}^{3}$ is the ground-truth translation vector. The translation vector and the rotation matrix are defined as:
\begin{equation}
    T_t^\ast =
    \begin{bmatrix}
        x_t^\ast \\
        y_t^\ast \\
        z_t^\ast
    \end{bmatrix},
    \quad
    R_t^\ast =
    \begin{bmatrix}
        r_{11}^\ast & r_{12}^\ast & r_{13}^\ast \\
        r_{21}^\ast & r_{22}^\ast & r_{23}^\ast \\
        r_{31}^\ast & r_{32}^\ast & r_{33}^\ast
    \end{bmatrix},
\end{equation}
where $x_t^\ast$, $y_t^\ast$, and $z_t^\ast$ specify the target position, and $\{r_{mn}^\ast\}_{m,n=1}^{3}$ define the rotation from the brick's intrinsic coordinate system $\mathcal{C}_{b_t}$ to the target-object coordinate system $\mathcal{C}$. Given the task prompt $P_{\mathrm{2}}$, the manual context $M_t$, the current assembly observation $S_t$, and the selected-brick geometry observation $b_t$, the MLLM parameterized by $\theta$ predicts the absolute target pose:
\begin{equation}
    \hat{P}_t =
    \begin{bmatrix}
        \hat{R}_t & \hat{T}_t \\
        \mathbf{0}^{\top} & 1
    \end{bmatrix}
    =
    \mathrm{MLLM}_{\theta}(P_{\mathrm{2}}, M_t, S_t, b_t).
\end{equation}

This task captures a realistic spatial reasoning requirement in assembly: predicting where and how a selected component should be placed within the target-object coordinate system. The formulation provides a unified, model-agnostic interface that is interpretable, directly evaluable in simulation, and compatible with downstream robotic execution.


\begin{figure*}[t]
    \centering
    \includegraphics[width=1.0\textwidth]{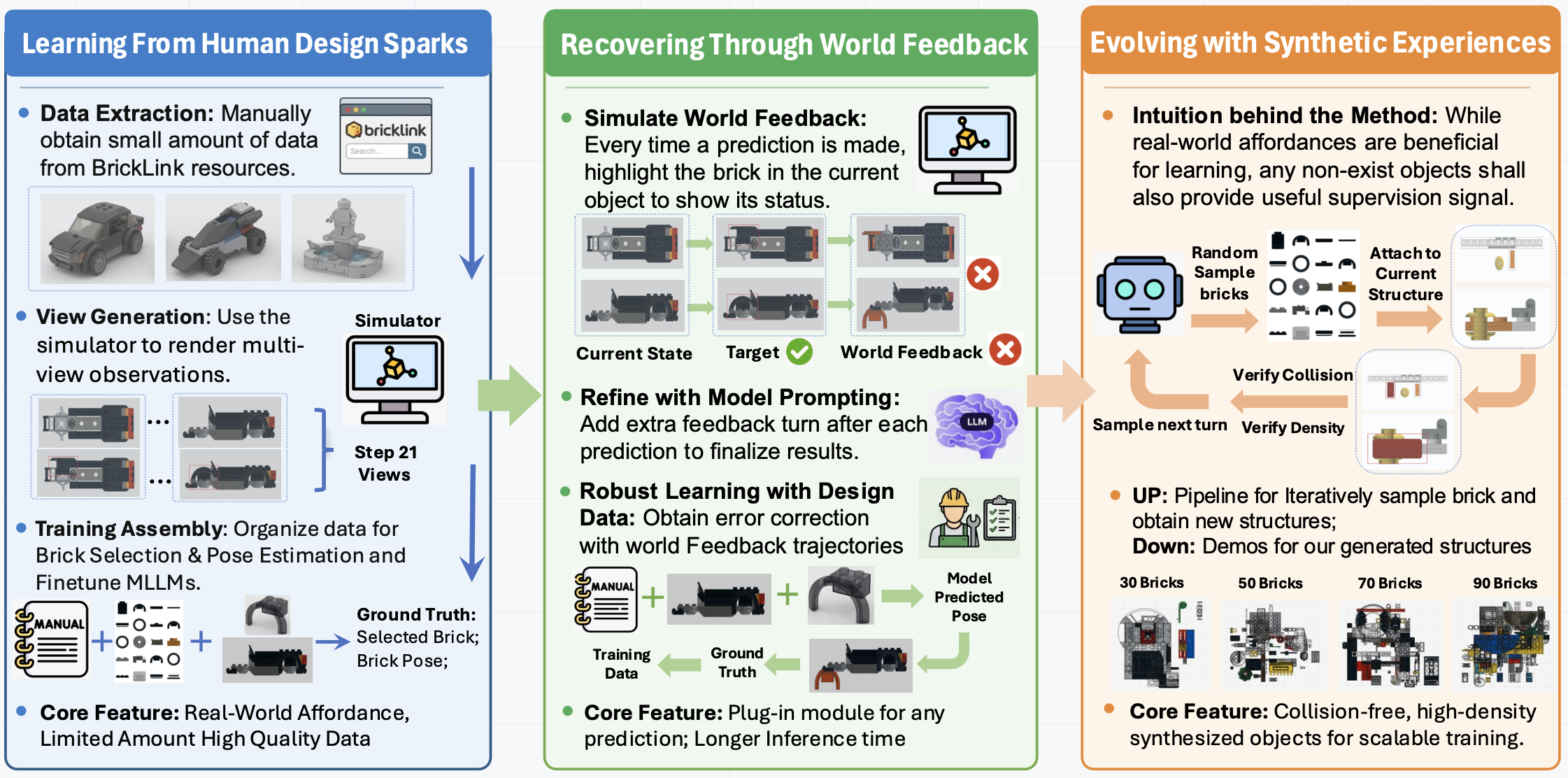}
    \caption{
Overview of the Brick-Composer learning framework. We improve assembly reasoning through three complementary signals: Human design supervision, World feedback for error recovery, and Scalable synthetic objects for experience expansion. Together, they substantially enhance the model's assembly capabilities.
}
    \label{fig:method}
\end{figure*}

\section{Methods}

\paragraph{Direct Prompting.}
We first evaluate whether general-purpose MLLMs can perform assembly through multimodal prompting. At each step, the model receives visual inputs for brick selection or pose estimation, together with task instructions and reasoning guidance. The complete prompt is provided in Appendix~\ref{app:prompt}. As shown in Table~\ref{tab:main_eval} and Section~\ref{sec:main results}, prompting alone remains insufficient for reliable brick selection and precise pose prediction, motivating our learning framework with \textit{Human Design Sparks}, \textit{World Feedback}, and \textit{Synthetic Experiences} shown in Figure~\ref{fig:method}.

\subsection{Learning from Human Design Sparks}

To move beyond zero-shot prompting, we adapt MLLMs with supervised data from human-designed brick assemblies. We call these data \textit{Human Design Sparks} because they capture not only valid placements, but also physically meaningful structures, feasible connections, creative construction patterns, and final objects with real-world affordances. Such signals are difficult to obtain from random configurations alone and help the model connect its real-world priors with concrete assembly decisions. We use our simulator to convert symbolic annotations into MLLM-ready visual inputs: six-view orthogonal before-and-after manuals with the current brick highlighted, and coordinate-augmented renderings of the current state and selected brick for spatial prediction.

We train both brick selection and pose estimation by serializing task outputs as natural-language answers and applying a standard autoregressive next-token objective. Each assembly step is treated as an independent training instance. For step $t$, let $V_t$ be the visual input, $I$ the task instruction, and $A_t=\{a_1,\ldots,a_K\}$ the answer token sequence. The objective is
\begin{equation}
\mathcal{L}_{\mathrm{Designer}}
=
-
\sum_{k=1}^{K}
\log p_{\theta}
\left(
a_k \mid a_{<k}, I, V_t
\right).
\end{equation}
The target answer is $A_t^{\mathrm{sel}} = y^* = (i^*, j^*)$ for brick selection and $A_t^{\mathrm{pose}} = P^* = (R^*, T^*)$ for pose estimation, where $R^* \in SO(3)$ and $T^* \in \mathbb{R}^3$. Thus, both tasks are optimized under the same autoregressive prediction framework.

\subsection{Learning from World Feedback}

Human assembly is rarely completed through a single prediction; constructors place a part, inspect the resulting structure, and correct mistakes. We introduce a similar world-feedback mechanism for MLLMs. As shown in Figure~\ref{fig:method}, our simulator approximates a physical assembly environment by executing the model's predicted selection or pose, rendering the resulting state, and highlighting the selected brick and placement. The model can therefore observe the consequence of its action and use visual discrepancies as a feedback.

Formally, at step $t$, the model receives visual input $V_t$ and instruction $I_t$, generates an initial response $\hat{A}_t$, and obtains a simulator-rendered feedback observation $O_t$. We use feedback in two settings. For prompting, we append $O_t$ to the trajectory and ask the model to revise its prediction:
\begin{equation}
\hat{A}_t^{\mathrm{revise}} =
\mathrm{MLLM}_{\theta}
\left(
I_t, V_t, \hat{A}_t, O_t
\right).
\end{equation}
For training, we construct correction trajectories from model errors and fine-tune the model to predict the corrected answer:
\begin{equation}
\mathcal{L}_{\mathrm{Fb}}
=-
\sum_{k=1}^{K}
\log p_{\theta}
\left(
a_{k}
\mid
a_{<k}, 
I_t,
V_t,
\hat{A}_t,
O_t
\right)
\end{equation}
This encourages the model to further learn to interpret world-state discrepancies and recover from wrong selections or inaccurate poses.

\subsection{Scaling with Synthetic Experiences}

Although human design provide high-quality supervision, their scale is limited by the availability of human-designed models, object-specific semantics, and copyright constraints, making brick selection and especially pose estimation difficult to learn at scale. Since both tasks mainly rely on fine-grained visual grounding, part geometry, and spatial reasoning, we further scale learning with \textit{Synthetic Experiences}: arbitrary but physically plausible configurations that expose the model to diverse brick placements and pose patterns beyond existing object designs. While these configurations lack real-world semantics and object-level affordances, they serve as a scalable source of assembly experience, enabling the model to practice a much broader range of local connections and spatial transformations.

We generate each configuration by sampling a new brick, attaching it to an existing partial structure through feasible connection points, and filtering placements by collision avoidance and structural connectivity. A density-based reward further encourages compact structures rather than sparse random layouts. Each accepted placement is treated as a synthetic assembly step, rendered in the same before-and-after visual format as designer supervision, and used for both brick selection and pose estimation. We scale this process to generate about 700 synthetic objects, each containing 20--90 assembly steps, yielding over 40K training steps and substantially expanding the supervision available for learning. We use the same training objective as designer supervision. For each synthetic step $t$, the model receives the rendered visual input $V_t^{\mathrm{arb}}$, task instruction $I_t$, and target answer $A_t^{\mathrm{arb}}$:
\begin{equation}
\mathcal{L}_{\mathrm{Arb}}
=
-
\sum_{k=1}^{K}
\log p_{\theta}
\left(
a_k \mid a_{<k}, I_t, V_t^{\mathrm{arb}}
\right).
\end{equation}

\begin{table*}[t]
\centering
\caption{Main evaluation of state-of-the-art MLLMs on BC-Bench. We report overall average and best-object performance, with higher-is-better metrics marked by $\uparrow$ and lower-is-better metrics marked by $\downarrow$. The results show that current MLLMs remain limited in reliable brick selection and precise pose estimation.}
\label{tab:main_eval}
\setlength{\tabcolsep}{4.5pt}
\renewcommand{\arraystretch}{1.12}
\resizebox{\textwidth}{!}{
\begin{tabular}{lcccccccc}
\toprule
\multirow{2}{*}{\textbf{Eval Metrics $\rightarrow$}} 
& \multicolumn{4}{c}{\textbf{Overall Average}} 
& \multicolumn{4}{c}{\textbf{Best Object Performance}} \\
\cmidrule(lr){2-5} \cmidrule(lr){6-9}
  & \textbf{Selection} $\uparrow$ 
& \textbf{PE Trans.} $\downarrow$ 
& \textbf{PE Rot.} $\downarrow$ 
& \textbf{Step-Wise} $\uparrow$
& \textbf{Selection} $\uparrow$ 
& \textbf{PE Trans.} $\downarrow$ 
& \textbf{PE Rot.} $\downarrow$ 
& \textbf{Step-Wise} $\uparrow$ \\
\multirow{1}{*}{\textbf{Model $\downarrow$ }} & \textbf{Acc. (\%)} 
& \textbf{Err(LDU)} 
& \textbf{Err. ($^\circ$)} 
& \textbf{SR (\%)}
& \textbf{Acc. (\%)} 
& \textbf{Err(LDU)} 
& \textbf{Err. ($^\circ$)} 
& \textbf{SR (\%)} \\
\midrule

Gemma-3-12B
& 4.35 
& 269.09 
& 63.00 
& 0.09 
& 17.24 
& \underline{41.86}
& \textbf{12.86}
& \underline{2.22} \\

InternVL-3.5-8B 
& 13.24 
& 221.69 
& 88.43 
& 0.00 
& 31.25 
& \textbf{41.54}
& 35.00
& 0.00 \\

Qwen-3-VL-8B 
& 22.76 
& \textbf{210.14}
& \underline{62.47}
& \textbf{0.36}
& 55.17
& 43.81
& \textbf{12.86}
& \textbf{4.44} \\

Qwen-3.5-VL-27B 
& \underline{37.44}
& 314.32 
& 82.94 
& \underline{0.18}
& \underline{75.86}
& 64.47
& \underline{34.84}
& \textbf{4.44} \\

GPT-5.4 
& \textbf{43.88}
& 310.78 
& 74.85 
& \underline{0.18}
& \textbf{93.10}
& 67.94
& 40.00
& \underline{2.22} \\

\bottomrule
\end{tabular}
}
\end{table*}

\section{Experiments}
\label{sec:experiments}


\begin{table*}[t]
\centering
\caption{Performance improvements on BC-Bench. Our learning framework consistently improves brick selection accuracy, reduces pose estimation errors, and increases step-level assembly success. World Feedback (P) denotes direct prompting with simulator feedback, while World Feedback (L) denotes learning from simulator-generated feedback trajectories jointly with human supervision data. Brick-Composer integrates all the learning signals.}
\label{tab:main_method}
\setlength{\tabcolsep}{4.5pt}
\renewcommand{\arraystretch}{1.12}
\resizebox{\textwidth}{!}{
\begin{tabular}{lcccccccc}
\toprule
\multirow{2}{*}{\textbf{Eval Metrics $\rightarrow$}} 
& \multicolumn{4}{c}{\textbf{Overall Average}} 
& \multicolumn{4}{c}{\textbf{Best Object Performance}} \\
\cmidrule(lr){2-5} \cmidrule(lr){6-9}
  & \textbf{Selection} $\uparrow$ 
& \textbf{PE Trans.} $\downarrow$ 
& \textbf{PE Rot.} $\downarrow$ 
& \textbf{Step-Wise} $\uparrow$
& \textbf{Selection} $\uparrow$ 
& \textbf{PE Trans.} $\downarrow$ 
& \textbf{PE Rot.} $\downarrow$ 
& \textbf{Step-Wise} $\uparrow$ \\
\multirow{1}{*}{\textbf{Approaches $\downarrow$ }} & \textbf{Acc. (\%)} 
& \textbf{Err(LDU)} 
& \textbf{Err. ($^\circ$)} 
& \textbf{SR (\%)}
& \textbf{Acc. (\%)} 
& \textbf{Err(LDU)} 
& \textbf{Err. ($^\circ$)} 
& \textbf{SR (\%)} \\
\midrule
\rowcolor{gray!15}
    \multicolumn{9}{c}{\textbf{Performance Comparison of Learning Approaches for Model: Gemma-3-12B}} \\ 

\textbf{Direct Prompting}
& 4.35 
& 269.09 
& 63.00 
& 0.09 
& 17.24 
& 41.86
& \textbf{12.86}
& 2.22 \\

\textbf{World Feedback (P)}
& --
& 273.41
& 62.26
& 0.00
& --
& 39.46
& \underline{14.53}
& 2.22 \\

\textbf{Designer Supervision}
& \underline{15.59}
& 201.51
& 55.69
& 0.72
& \underline{26.67}
& 36.91
& 15.00
& 4.44 \\

\textbf{World Feedback (L)}
& --
& \underline{170.33}
& \underline{51.49}
& \underline{1.99}
& --
& \underline{27.56}
& \textbf{12.86}
& \underline{9.07} \\

\textbf{Brick-Composer}
& \textbf{17.95}
& \textbf{123.69}
& \textbf{45.43}
& \textbf{4.35}
& \textbf{52.87}
& \textbf{24.63}
& \textbf{12.86}
& \textbf{18.75} \\

\rowcolor{gray!15}
    \multicolumn{9}{c}{\textbf{Performance Comparison of Learning Approaches for Model: Qwen-3-8B-VL}} \\

\textbf{Direct Prompting}
& 22.76 
& 210.14
& 62.47
& 0.36
& 55.17
& 43.81
& 12.86
& 4.44 \\

\textbf{World Feedback (P)}
& --
& 226.33
& 65.66
& 0.27
& --
& 42.36
& 12.86
& 4.44 \\

\textbf{Designer Supervision}
& \underline{48.29}
& 162.82
& 57.81
& 5.40
& \underline{73.24}
& 27.13
& 12.95
& 8.92 \\

\textbf{World Feedback (L)}
& --
& \underline{137.26}
& \underline{52.65}
& \underline{6.24}
& --
& \underline{21.46}
& \underline{11.64}
& \underline{15.36} \\

\textbf{Brick-Composer}
& \textbf{68.21}
& \textbf{65.63}
& \textbf{37.97}
& \textbf{14.27}
& \textbf{90.65}
& \textbf{14.29}
& \textbf{0.00}
& \textbf{41.63} \\

\bottomrule
\end{tabular}
}
\end{table*}

\subsection{Experiment Settings}

We evaluate state-of-the-art MLLMs from different model families and scales on BC-Bench, including Gemma-3-12B~\citep{gemmateam2025gemma3technicalreport}, InternVL3.5-8B~\citep{wang2025internvl35advancingopensourcemultimodal}, Qwen3-VL-8B~\citep{bai2025qwen3vltechnicalreport}, Qwen3.5-27B~\citep{qwen2026qwen35}, and GPT-5.4~\citep{openai2026gpt54}. We further apply our learning framework to Gemma-3-12B and Qwen3-VL-8B as representative open-weight models. 


\subsection{Evaluation Metrics}


\paragraph{Brick Selection Metrics} We report \textit{\textbf{Set-based Accuracy}} to handle physically identical and interchangeable candidate bricks. At time step $t$, the model predicts $(\hat{b}_t^i,\hat{b}_t^j)$, which is counted as correct if it belongs to the valid equivalent set $\mathcal{B}_t^*=\{(b^i,b^j)\in\mathcal{C}_t \mid \phi(b^i,b^j)=\phi(b_t^i,b_t^j)\}$, where $\mathcal{C}_t$ is the candidate set and $\phi(\cdot)$ maps each brick to equivalence-defining attributes such as part ID, color, and geometry. The brick selection accuracy is then defined as $\text{Acc}_{\text{brick}}=\frac{1}{N}\sum_{t=1}^{N}\mathbb{I}\big((\hat{b}_t^i,\hat{b}_t^j)\in\mathcal{B}_t^*\big)$, where $N$ is the number of evaluated steps.

\paragraph{Brick Pose Estimation Metrics}
We evaluate the predicted absolute pose of the selected brick in the shared target-object coordinate system. At step $t$, let the predicted pose be $\hat{P}_t=(\hat{R}_t,\hat{T}_t)$ and the ground-truth pose be $P_t=(R_t,T_t)$. We measure \textit{\textbf{Mean Translation Error}} as $e^{\text{trans}}=\frac{1}{N}\sum_{t=1}^{N}\|\hat{T}_t-T_t\|_2$. We measure \textit{\textbf{Mean Rotation Error}} by the angular distance between rotations, where $\theta_t=\cos^{-1}\left(\frac{\mathrm{Tr}(R_t^\top\hat{R}_t)-1}{2}\right)$ and $e^{\text{rot}}=\frac{1}{N}\sum_{t=1}^{N}\frac{180}{\pi}\theta_t$. Since many bricks have rotational symmetries, we compute rotation error in a symmetry-aware manner. Let $\mathcal{S}_t$ be the set of valid symmetry rotations for the brick at step $t$. We define $\theta_t^{\text{sym}}=\min_{S\in\mathcal{S}_t}\cos^{-1}\left(\frac{\mathrm{Tr}((R_tS)^\top\hat{R}_t)-1}{2}\right)$ and report the final mean rotation error as $e^{\text{rot}}=\frac{1}{N}\sum_{t=1}^{N}\frac{180}{\pi}\theta_t^{\text{sym}}$. Accordingly, orientations that differ only by a valid brick symmetry are not penalized.

\paragraph{Joint Assembly Evaluation Metrics}
We further evaluate joint performance using \textit{\textbf{Averaged Step-level Success Rate}}, where a step is successful only if the model selects the correct brick and predicts the target pose exactly within our simulated environment. Although practical systems may use collision handling to correct small pose deviations, we adopt a strict zero-tolerance criterion. We report all metrics from two perspectives: \textit{overall average}, computed across the full test set, and \textit{best-object performance}, computed on the single object where each model performs best.

\subsection{Main Results}
\label{sec:main results}

\begin{figure}[t]
    \centering
    \includegraphics[width=0.95\linewidth]{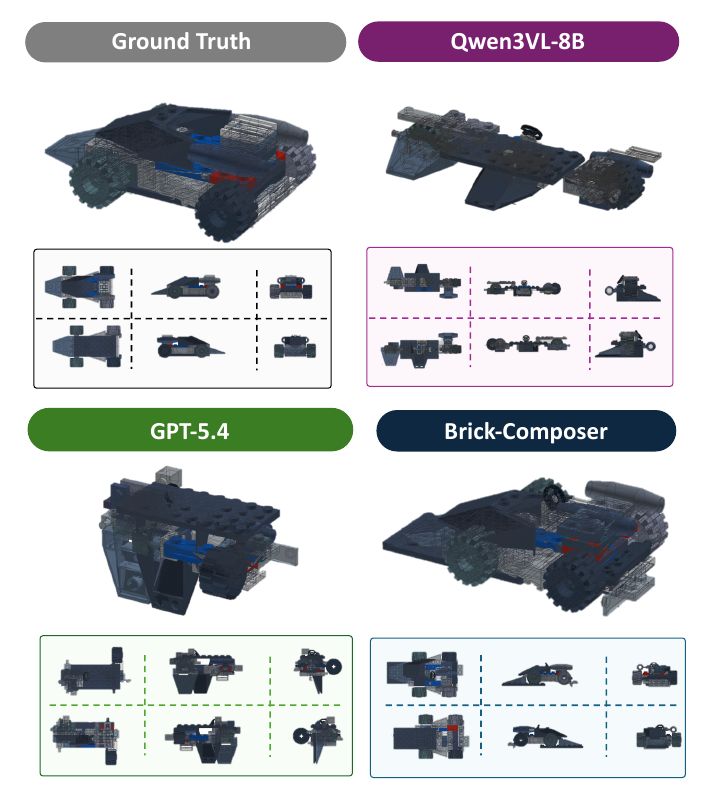}
    \vspace{-0.1in}
    \caption{Qualitative Examples of Model Assembly}
    \vspace{-0.2in}
    \label{fig:case}
\end{figure}

\paragraph{Limitation of General-Purpose MLLMs}
Table~\ref{tab:main_eval} shows that general MLLMs have limited zero-shot assembly capability. For brick selection, stronger models achieve non-trivial but still unreliable performance: GPT-5.4 reaches the best overall accuracy, followed by Qwen-3.5-VL-27B. This suggests that current MLLMs can partially ground the target brick from visual context. The gap is larger for pose estimation: most models produce very large translation and rotation errors, indicating that these models lack robust 3D spatial reasoning for precise placement. As a result, strict step-level success remains near zero, with the best overall success rate only 0.36\% and the best-object success rate only 4.44\%. These results show that prompting alone is insufficient for reliable assembly. More analyses are in Appendix~\ref{app:exp_details}.

\paragraph{Brick-Composer Learning Performance}
As shown in Table~\ref{tab:main_eval}, the final Brick-Composer framework substantially improves both Gemma-3-12B and Qwen-3-VL-8B. Both models improve brick selection accuracy, greatly reduce translation and rotation errors, and achieve much higher step-wise success rates. On the best-object evaluation, the fine-tuned Qwen-3-VL-8B achieves near-perfect rotation prediction and a step-wise success rate of $42\%$. Its average translation error of $14.29$ LDU is within one LEGO stud in most cases, indicating strong placement accuracy. Figure~\ref{fig:case} shows a qualitative example where the assembled object visually resembles the target structure.

\subsection{Ablation Studies}

We conduct ablation studies to verify the effectiveness of each learning signal in our framework. As shown in Table~\ref{tab:main_method}, learning from designer supervision alone improves model performance despite limited data, especially for brick selection, while also reducing translation and rotation errors. In contrast, applying world feedback alone brings limited gains and can even hurt performance, as base models are not yet familiar with pose estimation and the additional feedback views introduce extra visual reasoning challenges. However, when combined with designer supervision, world feedback becomes effective and further improves upon the designer-supervised model. This benefit comes with additional inference cost: world feedback requires rendering erroneous states and prompting the model for another round of reasoning at each step. For a $70$-step object, this can add roughly one hour of inference time. Finally, scaling with synthetic experience provides another substantial performance boost, pushing assembly capability to a new level. We leave a systematic study of scaling trends with data and model size to future work.

\section{Conclusion}

We formulate MLLM-based brick assembly as a sequential decision problem consisting of two coupled subtasks: brick selection and brick pose estimation. While our BC-Bench shows that general-purpose MLLMs remain limited in this setting, especially for precise pose prediction, we introduce Brick-Composer, a unified learning framework that adapts MLLMs through three complementary signals: human-designed assembly data, simulator-based world feedback, and scalable synthetic configurations. Our results show that Brick-Composer substantially improves assembly performance by increasing fine-grained brick selection accuracy, reducing translation and rotation errors, and raising strict step-level success. These results suggest that construction-oriented capabilities are learnable for large models when they are scaled with physically grounded supervision and feedback.

Looking forward, our data provides a challenging yet meaningful testbed for studying how AI agents can move beyond perception toward compositional spatial reasoning and executable action in the real world. We hope our solution also encourages future research on large-scale assembly learning, spatially grounded model improvement, and scalable agents that can transfer from simulated construction environments to real-world robotics, manufacturing, and interactive design assistance.

\section*{Limitations} 

Despite the well-constructed benchmark for evaluating MLLMs in assembly and the effectiveness of the Brick-Composer framework, this work has several limitations that also point to promising future directions. First, Brick-Composer studies brick assembly in simulation rather than direct real-world robotic execution. This choice allows us to isolate the core multimodal reasoning problem: whether MLLMs can understand assembly states, identify the correct component, and predict physically meaningful 3D placements under controlled and reproducible conditions. The simulator further provides accurate pose labels, consistent multi-view rendering, and scalable feedback, which are difficult to obtain at the same scale in real-world robotic settings. Nevertheless, transferring these capabilities to physical robots would require handling additional factors such as perception noise, occlusion, calibration errors, grasping constraints, contact dynamics, and execution failures. Thus, our results should be viewed as a strong step toward the spatial reasoning foundation needed for embodied assembly, rather than an end-to-end robotic assembly system.

Second, although human-designed assembly data provides realistic construction patterns and high-quality supervision, its scale is naturally constrained by the availability and copyright status of BrickLink-style design resources. This limitation motivates one of the key design choices of Brick-Composer: complementing human-designed data with simulator-generated arbitrary configurations. Such synthetic configurations allow us to scale physically valid supervision beyond the direct use of copyrighted designs, while still training the model on brick selection, connection geometry, and pose reasoning. At the same time, these configurations may not fully capture the semantic structure, aesthetic preference, and higher-level construction logic of designed objects. Future work could further explore copyright-aware data curation, procedurally generated object designs, and stronger methods for combining realistic human design priors with scalable synthetic experience.

Third, our evaluation focuses on step-wise assembly, where the model selects and places the next brick under a given assembly context. This setting is intentional: it provides a clear and measurable formulation for evaluating the two fundamental capabilities required by assembly, namely fine-grained brick selection and precise pose estimation. However, full object-level construction requires composing many such decisions over long horizons, where early mistakes can influence later steps. While Brick-Composer demonstrates that construction-oriented capabilities are learnable for MLLMs through physically grounded supervision and feedback, extending these gains to robust long-horizon autonomous construction remains an important direction for future work.

\section*{Ethical Statement on LLM Assistance}

We primarily use GPT-5 as a tool for language refinement, including polishing text and improving clarity. All model-generated content is thoroughly reviewed and rewritten by human authors to ensure accuracy, originality, and adherence to research integrity standards.

\bibliography{custom}

\newpage

\appendix
\begin{figure*}[t]
    \centering
    \includegraphics[width=1.0\textwidth]{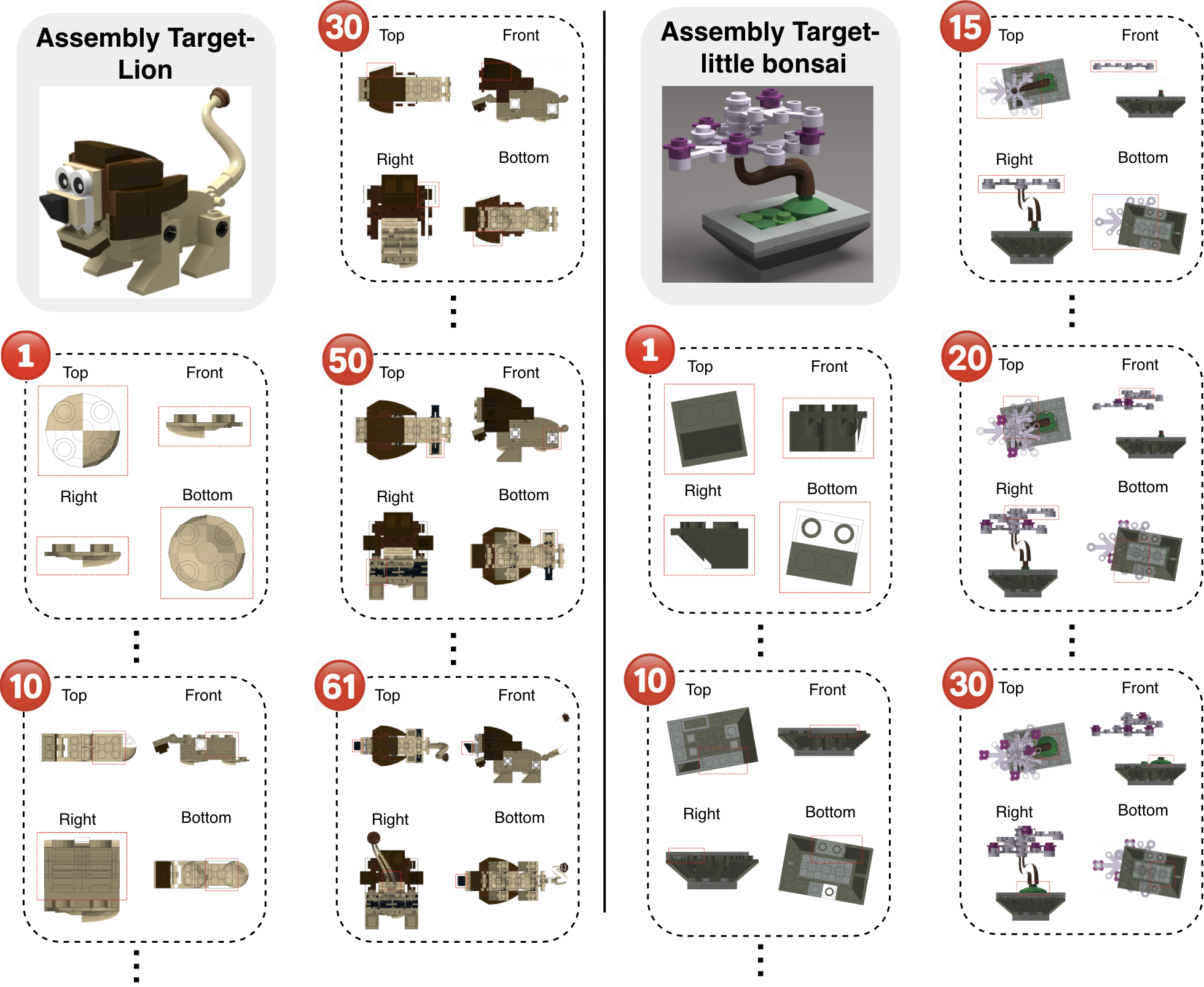}
    \caption{
    Examples of manual-style assembly sequences in BC-Bench. Each example shows a target LEGO-style object and selected construction steps rendered from multiple orthogonal views, with red boxes marking the newly added brick. BC-Bench provides six orthogonal views for each assembly step, along with annotations for brick identity and pose. The parts view data are shown in Figure~\ref{fig:assembly_part}.
    }
    \label{fig:assembly_examples}
\end{figure*}

\begin{figure*}[t]
    \centering
    \includegraphics[width=1.0\textwidth]{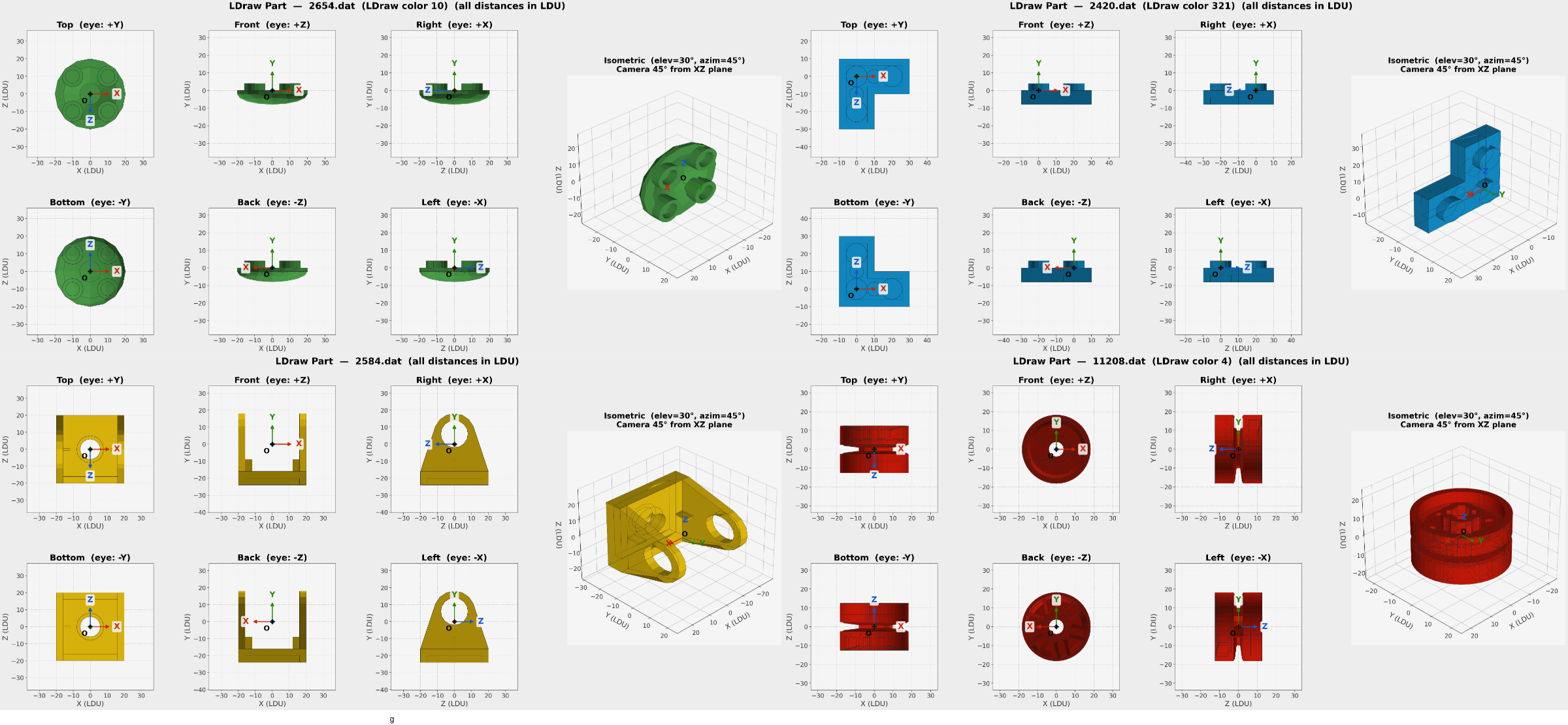}
    \caption{
    Examples of our rendered part-demo data in BC-Bench, we visualize the part within its own coordinate system, so that the model could have better knoweledge over their scale information.  
    }
    \label{fig:assembly_part}
\end{figure*}

\section{Dataset for MLLM Assembly}
\label{app:dataset}

\paragraph{Visualization}
Figure~\ref{fig:task} illustrates the overall data formulation of BC-Bench, including the manual context, candidate bricks, current assembly state, and target pose annotation. To further clarify the benchmark format, Figure~\ref{fig:assembly_examples} shows step-wise rendered assembly views, where each construction step is visualized from multiple viewpoints and the newly added brick is highlighted. Figure~\ref{fig:assembly_part} provides additional examples of part-level inputs, showing the diverse brick geometries, colors, and metadata used for brick selection and pose estimation. Together, these visualizations demonstrate how BC-Bench converts human-designed objects into structured step-wise assembly tasks for MLLMs.

\paragraph{Data Collection Details and Statistics}
To respect BrickLink designers' copyright, we do not conduct large-scale scraping or use large-scale designer data for training or evaluation. Instead, we manually downloaded $102$ object files for research purposes and built our own simulator to render multi-view assembly states with step-wise annotations. We split the data at the object level with a $0.8:0.2$ train-test ratio, rather than at the step level, to avoid placing highly similar steps from the same object in both splits. This ensures that evaluation measures object-level assembly generalization rather than memorization of individual construction steps. Overall, BC-Bench contains $3{,}873$ training steps from $82$ objects and $1{,}013$ evaluation steps from $20$ objects, covering a diverse set of LEGO-style objects and construction procedures.

\section{Trajectories for Multimodal Reasoning}
\label{app:prompt}

We provide representative model interaction trajectories to illustrate our prompting design and task formulation for three settings: (1) brick selection, (2) brick pose estimation, and (3) world-feedback prompting, where models learn from simulator-rendered feedback. For readability, we omit the original images in these trajectories and replace each image position with the placeholder \texttt{<Image>} following the ShareGPT data format.

\begin{PromptforBrickSelection}
\textbf{"system"}: "You are a LEGO brick assembly expert. You will be given three images:\newline 

1. Original views: 6 orthogonal views (Top, Front, Right, Bottom, Back, Left) of the current LEGO assembly BEFORE the assembly step.\newline 

2. Target views: the same 6 orthogonal views AFTER the brick is placed, with the newly added brick highlighted with a red dashed bounding box.\newline 

3. Remaining brick catalog: a grid showing every brick that still needs to be placed (including the current step's brick), with part filenames (e.g. 3023.dat) labeled above each tile. \newline 

\textbf{Your task:} Compare the Original and Target views to identify which brick was placed, then locate it in the catalog. Output format 2014 one line per brick:  \{part-filename\}, row \{R\}, col \{C\}\newline 

\textbf{Rules:} \{part-filename\} must be the exact .dat filename shown in the catalog label. \{R\} and \{C\} are the 1-based row and column in the catalog grid (6 columns wide). If multiple bricks are placed in this step, output one line per brick in the same order as requested. Do NOT output any explanation, preamble, or extra text  2014 only the formatted line(s).'',\newline

    ``conversations'': [
      \{
        "from": "human",
        "value": "Below are the three inputs for this assembly step.\newline
        
        Image 1  Original views (brick not yet placed, 6 orthogonal angles: Top, Front, Right, Bottom, Back, Left): <image>\newline
        
        Image 2 Target views (brick placed, same 6 angles; newly added brick is highlighted with a red dashed bounding box): <image>\newline
        
        Image 3 Remaining brick catalog (rows and columns are 1-indexed from top-left, 6 columns wide; part names are labeled above each tile):<image> This step places exactly ONE brick. \newline
        
        Output a single line in the format:  <part.dat>, row <R>, col <C>"
      \},
      {
        "from": "gpt",
        "value": "3035.dat, row 6, col 3"
      \}
    ],
}
\end{PromptforBrickSelection}

\begin{PromptforBrickPoseEstimation}
{
    \textbf{"system"}: "You are a LEGO brick assembly expert with precise knowledge of 3D coordinate systems and LDraw file formats. You will be given:  \newline
    
    1. Original views: a composite of 6 orthogonal views (Top, Front, Right, Bottom, Back, Left) of the current LEGO assembly BEFORE the assembly step.\newline
    
    2. Target views: the same 6 orthogonal views of the assembly AFTER the brick is placed.  The newly added brick is highlighted with a red dashed bounding box.\newline
    
    3. Part render(s): a rendered image of the brick (or bricks) to be placed, with the LDraw local coordinate frame annotated (red=+X, green=+Y, blue dot=+Z).\newline
    
    4. Current-state axes render: a 7-view composite (6 orthogonal + 1 isometric) of the assembly BEFORE this step, with LDU coordinate tick labels on every view and an isometric 3D projection in the right column.  Use this to read off exact LDU coordinates and understand the spatial layout.\newline
    
    You will also be given text with part metadata (name, approximate physical size).\newline
    
    \textbf{Your task:} Determine the absolute global position (x, y, z in LDraw Units, LDU) and the rotation matrix that transforms the part's local frame into the global frame, for each brick being placed in this step. Coordinate system conventions (LDraw global frame): X axis  : positive goes RIGHT in the standard front view.  Y axis  : positive goes DOWN  (LDraw uses a left-handed, Y-down system). Z axis  : positive goes TOWARD the viewer in the standard front view. 1 LDU   = 0.4 mm  (one standard LEGO stud pitch = 20 LDU = 8 mm). Rotation matrix convention:  The 300d73 matrix R encodes the orientation of the part.  Each row gives the components of the brick's local axis in world space:\newline
    
    row 0 = brick local X axis expressed in world coordinates \newline
    row 1 = brick local Y axis expressed in world coordinates\newline  
    row 2 = brick local Z axis expressed in world coordinates\newline
    
    \textbf{Output format} one block per brick (numbered Brick 1, Brick 2, when multiple):  Brick <N>: Translation: x=<float>, y=<float>, z=<float>  Rotation:\newline
    
    [<r00>, <r01>, <r02>]\newline
    [<r10>, <r11>, <r12>]\newline
    [<r20>, <r21>, <r22>]\newline
    
    \textbf{Rules:} Use decimal floats (e.g. 30.0, -50.0001, 11.1001). If there is only one brick in the step, still use "Brick 1:" as the header.  Do NOT output any explanation, preamble, or extra text 2014 only the formatted blocks above.",\newline

    "conversations": [
      \{
        "from": "human",
        "value": "Below are all inputs for this assembly step. \newline
        
        Image 1 Original views (assembly BEFORE this step, 6 orthogonal angles: Top, Front, Right, Bottom, Back, Left): <image> \newline
        
    Image 2 Target views (assembly AFTER this step; the newly placed brick is highlighted with a red dashed bounding box): <image> \newline
    
    Image 3 Part render: The brick (3035.dat (color 0)) The local LDraw coordinate frame is annotated on the image: red arrow = +X (right), green arrow = +Y (down), blue dot = +Z (toward viewer). <image>\newline
    
    Image 4 Current-state axes render (7 views: Top/Front/Right/Bottom/Back/Left + Isometric; light-grey background with LDU coordinate tick labels; use this to read off spatial positions and understand the 3D layout):<image>\newline
    
    Part information (brick to place this step):\newline
    
    Brick 1: 3035.dat  |  3035  |  dims 2248 3.2 x 6.4 x 0.51 cm  |  LDraw color 0 \newline
    
    Using all the images and part information above, predict the absolute global position (x, y, z in LDU) and rotation matrix for each of the brick listed above. Output exactly one 'Brick N:' block per brick, in the same order as listed."
      \},
      \{
        "from": "gpt",
        "value": "Brick 1: \newline
        
        Translation:\newline
        x=0.0, y=-8.0, z=0.0 \newline
        
        Rotation:\newline
        [1.0, 0.0, 0.0] \newline
        [0.0, 1.0, 0.0]\newline
        [0.0, 0.0, 1.0]"\newline
      \}
    ],
}
\end{PromptforBrickPoseEstimation}

\begin{PromptforLearningwithWorldFeedback}
{
    \textbf{"system":} "You are a LEGO brick assembly expert with precise knowledge of 3D coordinate systems and LDraw file formats. You will be given:\newline
    
    1. Original views: a composite of 6 orthogonal views (Back, Bottom, Left, Right, Up, Front) of the current LEGO assembly BEFORE the assembly step.\newline
    
    2. Target views: the same 6 orthogonal views of the assembly AFTER the brick is placed.  The newly added brick is highlighted with a bounding box.\newline
    
    3. Part render(s): a rendered image of the brick (or bricks) to be placed, with the LDraw local coordinate frame annotated on the image.\newline
    
    4. Previous assembly render (omitted for the very first step): a composite rendering of the entire assembly from the immediately preceding step, giving you spatial context about the accumulated model.\newline
    
    5. Erroneous prediction render (when available): a rendering in which the incorrectly placed brick is shown in red, produced from a previous model prediction.  Accompanying text specifies the wrong predicted rotation matrix and scalar error magnitudes (Euclidean translation error and geodesic rotation error) only — no absolute translation values or per-axis offsets are given. Treat this as a negative example — analyse where and how the prior prediction went wrong visually, then use that reasoning to arrive at the correct pose. You will also be given text with part metadata (name, approximate physical size).\newline
    
    \textbf{Your task:} Determine the  absolute global position (x, y, z in LDraw Units, LDU) and the 3×3 rotation matrix** that transforms the part's local frame into the global frame, for each brick being placed in this step. Coordinate system conventions (LDraw global frame):  \newline
    
    • X axis  : positive goes RIGHT in the standard front view. \newline
    • Y axis  : positive goes DOWN  (LDraw uses a left-handed, Y-down system).\newline
    • Z axis  : positive goes TOWARD the viewer in the standard front view.\newline
    • 1 LDU   = 0.4 mm  (one standard LEGO stud pitch = 20 LDU = 8 mm). Rotation matrix convention:  The 3×3 matrix R encodes the orientation of the part. \newline
    
    Each column gives the direction of one of the part's local axes expressed in the global frame:    column 0 = direction of part's local X axis in global coordinates    column 1 = direction of part's local Y axis in global coordinates    column 2 = direction of part's local Z axis in global coordinates\newline
    
    Output format — one block per brick (numbered Brick 1, Brick 2, … when multiple): Brick <N>:  \newline
    
    Translation:\newline
    x=<float>, y=<float>, z=<float>  \newline
    
    Rotation:  \newline
    [<r00>, <r01>, <r02>]  \newline
    [<r10>, <r11>, <r12>]\newline
    [<r20>, <r21>, <r22>]\newline
    
    \textbf{Rules:}  \newline
    
    • Use decimal floats (e.g. 30.0, -50.0001, 11.1001).\newline
    • If there is only one brick in the step, still use "Brick 1:" as the header.\newline
    • Do NOT output any explanation, preamble, or extra text — only the formatted blocks above.",
    "conversations": [
      \{
        "from": "human",
        "value": "Below are all inputs for this assembly step.\newline
        
        Image 1 — Original views (assembly BEFORE this step, 6 orthogonal angles: Back, Bottom, Left, Right, Up, Front): <image>\newline
        
        Image 2 — Target views (assembly AFTER this step; the newly placed brick is highlighted with a bounding box): <image>\newline
        
        Image 3 — Part render: The brick (3023.dat — (color 71)) The local LDraw coordinate frame is annotated on the image: red arrow = +X (right), green arrow = +Y (down), blue dot = +Z (toward viewer)<image>\newline
        
        Image 4 — ERRONEOUS prior prediction (brick shown in RED — this is WRONG, do NOT copy it). A previous model attempt predicted the following incorrect pose for the brick(s) in this step.  The image renders the brick(s) in red at the wrongly predicted location/orientation so you can see visually where the mistake is.  Study the error and reason toward the correct answer. \newline
        
        Wrong prediction details: Brick 1 (3023.dat):  Wrong predicted rotation matrix (row-major):\newline
        [1.0000, 0.0000, 0.0000]\newline
        [0.0000, 1.0000, 0.0000]\newline
        [0.0000, 0.0000, 1.0000]\newline
        
        Euclidean translation error  : 23.7487 LDU  Geodesic rotation error      : 90.0000° <image>\newline
        
        Part information (brick to place this step): Brick 1: 3023.dat  |  3023  |  dims = 0.8 x 1.6 x 0.51 cm  |  LDraw color 71. Using all the images and part information above, predict the absolute global position (x, y, z in LDU) and 3 x 3 rotation matrix for each of the brick listed above. \newline
        
        Output exactly one 'Brick N:' block per brick, in the same order as listed."
      \},
      \{
        "from": "gpt",
        "value": "Brick 1:\newline
        
        Translation: \newline
        x=-10.0, y=-8.0, z=-20.0 \newline 
        
        Rotation: \newline
        [0.0, 0.0, -1.000001] \newline
        [0.0, 1.0, 0.0] \newline
        [1.000001, 0.0, 0.0]" \newline
      \}
    ]
}
\end{PromptforLearningwithWorldFeedback}

\begin{figure*}
    \centering
    \includegraphics[width=0.95\linewidth]{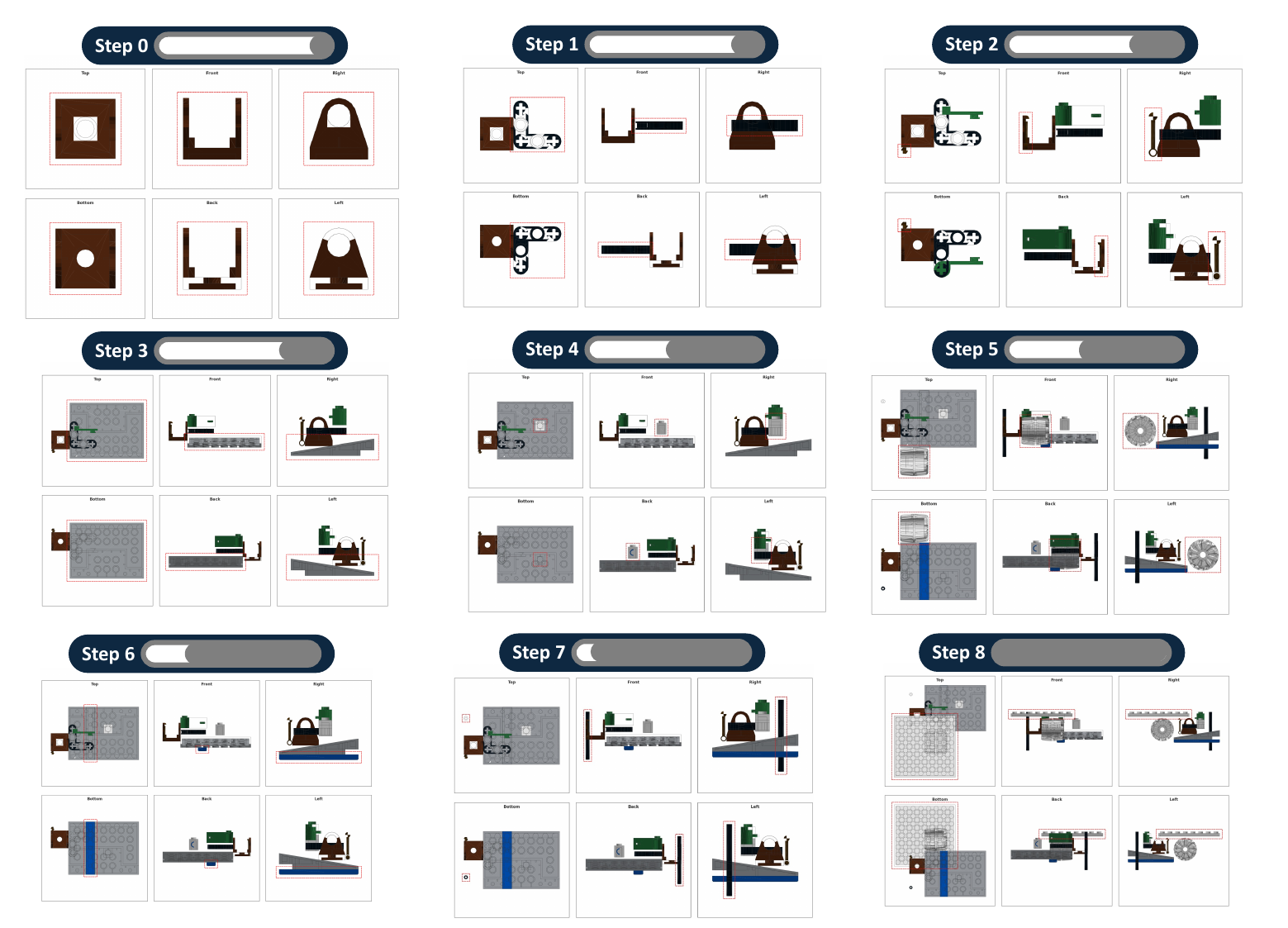}
    \caption{
    Examples of synthesized assembly configurations used for synthetic experience learning. Each structure is generated by incrementally attaching sampled bricks through feasible connection positions, while filtering invalid placements based on collision avoidance and structural connectivity. A density preference further encourages compact layouts, allowing the synthesized data to provide diverse and scalable supervision for brick selection and pose estimation.
    }
    \label{fig:synthetic_examples}
\end{figure*}

\begin{figure*}
    \centering
    \includegraphics[width=0.95\linewidth]{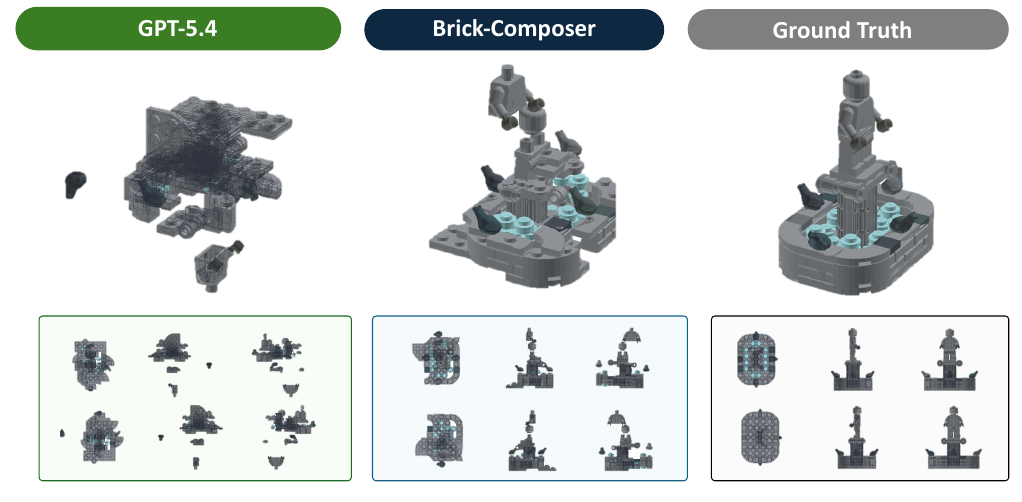}
    \caption{
    Additional qualitative examples of Brick-Composer assembly results. The generated assemblies show that the fine-tuned model can recover the overall structure and major components of target LEGO-style objects across multiple construction steps. These examples demonstrate the model's improved ability to jointly select appropriate bricks and place them into coherent object-level assemblies.
    }
    \label{fig:more_case_1}
\end{figure*}

\begin{figure*}
    \centering
    \includegraphics[width=0.95\linewidth]{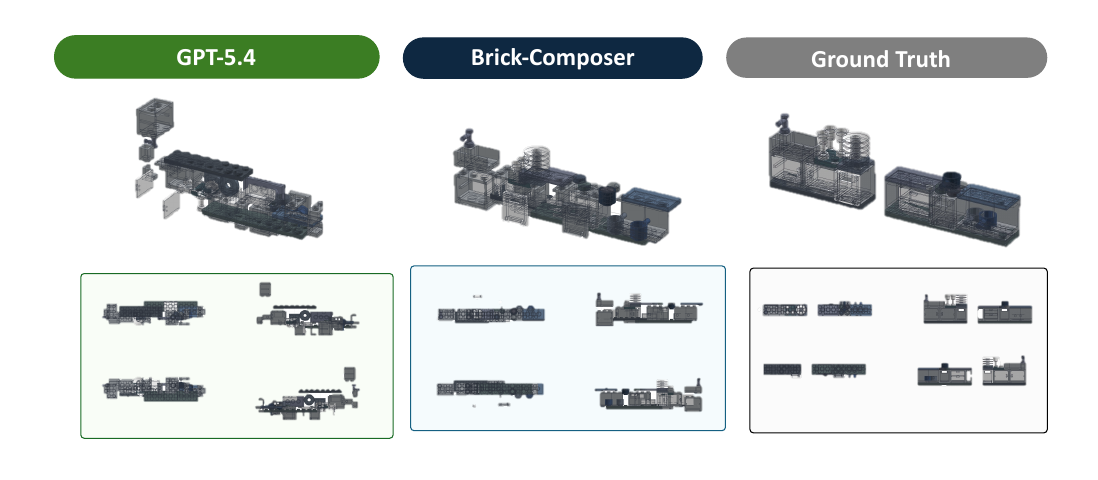}
    \caption{
    Additional qualitative examples of Brick-Composer assembly results. The generated assemblies show that the fine-tuned model can recover the overall structure and major components of target LEGO-style objects across multiple construction steps. These examples demonstrate the model's improved ability to jointly select appropriate bricks and place them into coherent object-level assemblies.
    }
    \label{fig:more_case_2}
\end{figure*}

\begin{figure*}
    \centering
    \includegraphics[width=0.95\linewidth]{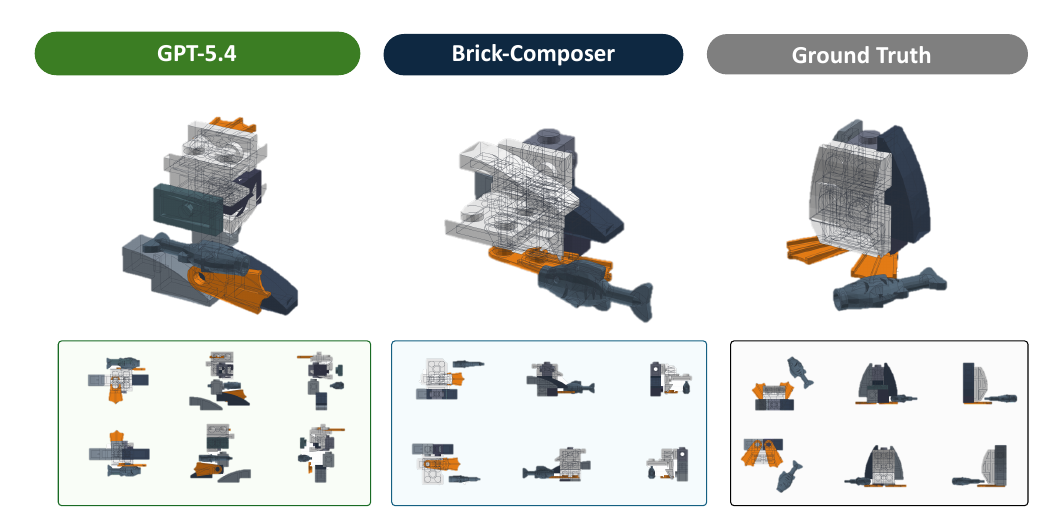}
    \caption{
    Additional qualitative examples of Brick-Composer assembly results.
    }
    \label{fig:more_case_3}
\end{figure*}

\begin{figure*}
    \centering
    \includegraphics[width=0.95\linewidth]{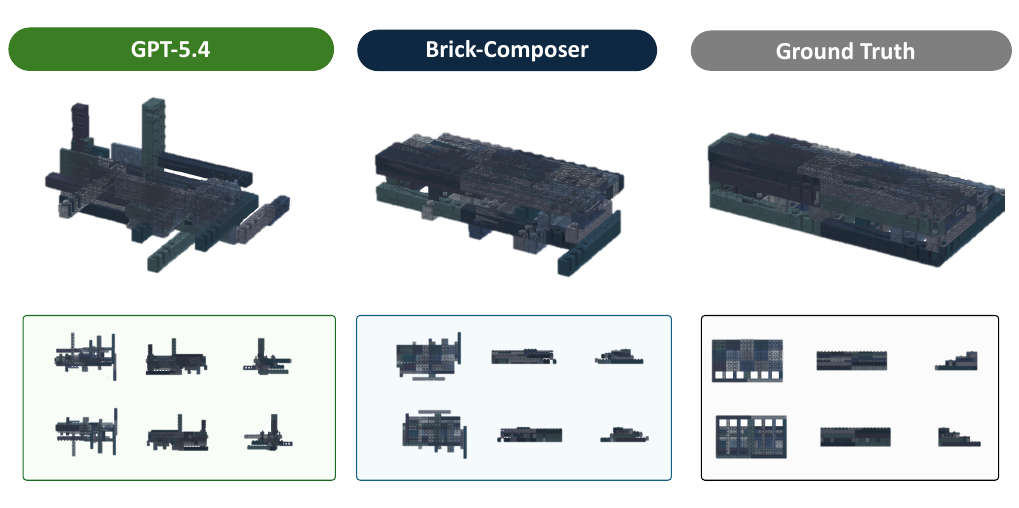}
    \caption{
    Additional qualitative examples of Brick-Composer assembly results.
    }
    \label{fig:more_case_4}
\end{figure*}

\section{More Analysis and Demos}
\label{app:exp_details}

\paragraph{Step-Wise Brick Selection Complexity}
We further analyze how brick selection difficulty changes across assembly steps. Our results show that selection accuracy is generally lower in the early stage of assembly, especially within the first $80\%$ of steps, while the last $20\%$ of steps tend to be easier. This pattern suggests that early-stage assembly is more ambiguous: the partial structure is still sparse, the manual provides limited accumulated context, and many candidate bricks may appear visually similar or functionally interchangeable. In contrast, later steps often benefit from a more complete object structure, where the target placement region and required brick type become easier to infer from the accumulated assembly context. This finding reveals a critical challenge for real-world assembly agents. Errors made in early brick selection are not isolated; selecting the wrong brick can change the subsequent structure, misalign later pose predictions, and trigger cascading failures throughout the remaining assembly process. Therefore, improving brick selection is not merely a preliminary localization problem, but a core requirement for reliable long-horizon assembly. In this sense, robust brick selection is as important as pose estimation, since accurate placement is only meaningful when the correct component has first been identified. 

\paragraph{Synthesized Examples}
We provide additional visualizations of the synthetic experience data used in our learning framework. As shown in Figure~\ref{fig:synthetic_examples}, these examples are not copied from human-designed objects, but are procedurally generated by sampling bricks and attaching them through feasible connection positions. During generation, each accepted structure must satisfy basic physical constraints, including collision avoidance and structural connectivity. We further apply a density preference to encourage compact and meaningful assemblies rather than sparse or randomly scattered layouts. These synthesized examples expand the diversity of training configurations and provide scalable supervision beyond the limited set of human-designed objects.

\paragraph{More Brick-Composer Qualitative Examples}
We show additional qualitative examples of Brick-Composer assembly results in Figure~\ref{fig:more_case_1}, Figure~\ref{fig:more_case_2}, Figure~\ref{fig:more_case_3}, and Figure~\ref{fig:more_case_4}. These examples illustrate how the learned model can compose multi-step LEGO-style structures that visually resemble the target objects. While some fine-grained placement errors may still remain, the overall shapes and major structural components are successfully recovered, showing that our learning framework enables MLLMs to acquire non-trivial assembly capability beyond direct prompting.

\section{Future Work}

\paragraph{Scaling Assembly Data and Spatial Skill Learning.}
Our results suggest that improving the spatial capabilities of MLLMs for brick assembly is a promising and scalable direction. A natural next step is to expand both human-designed and synthetically generated assembly data, while preserving key physical and procedural constraints such as collision avoidance, stable support, connector compatibility, and step-wise buildability. Prior work on brick generation has shown the value of large-scale buildable brick structures for learning stable design priors~\citep{pun2025brickgpt}, while recent graph-backed brick assembly work further highlights the importance of representing part types, connection semantics, and assembly sequences beyond simple voxel-like structures~\citep{kulits2026bricknet}. For Brick Composer, such scaling should not only increase data volume, but also increase the diversity of brick geometries, connection mechanisms, and instruction styles. This would allow future models to learn more robust spatial priors for selecting parts, estimating poses, and composing long-horizon assembly trajectories.

\paragraph{Internalizing Assembly and Design Knowledge.}
A second direction is to internalize reusable assembly and design knowledge into the model, rather than requiring the model to repeatedly infer all constraints from the prompt or demonstrations. This direction is inspired by recent work on policy internalization for LLM agents, where complex policy documents are parsed, categorized, and distilled into model behavior through continued pretraining or staged training~\citep{liu2025internalizing,wang2025mpi}. Although these works focus on conversational or policy-following agents, the core idea is highly relevant to assembly: many assembly rules are also structured, reusable, and difficult to follow purely through in-context prompting. For example, brick compatibility, collision constraints, symmetry handling, connector affordances, and design preferences can be treated as domain policies that guide the model's generation and verification process. During the dreaming process, such internalized knowledge could help generate more physically plausible and instruction-consistent synthetic assemblies. During the assembly process, it could help the model avoid infeasible placements, select parts with compatible affordances, and reason about local constraints without relying only on surface-level visual matching. This provides a practical bridge between data scaling and reliable spatial reasoning: instead of only collecting more examples, future systems can explicitly extract, organize, and internalize the rules that make those examples valid.

\paragraph{Digital Assembly with Computer-Use Agents.}
Another practical direction is to connect predicted assembly decisions with executable actions in digital design environments. Current Brick Composer focuses on understanding the correct brick and its target pose, but a deployable assembly assistant should eventually operate inside tools such as BrickLink Studio by issuing mouse, keyboard, and interface actions. This connects our task to recent computer-use agent benchmarks and methods, including OSWorld and macOSWorld, which evaluate agents in realistic desktop environments~\citep{xie2024osworld,yang2025macosworld}, and OSExpert, which studies how agents can learn reusable professional GUI skills through environment exploration~\citep{liu2026osexpert}. In our setting, the goal is not merely to automate arbitrary GUI actions, but to ground GUI operations in assembly semantics: selecting a part, rotating it, snapping it to a compatible connector, checking whether the placement is valid, and revising the design when conflicts occur. Recent work on reliable computer-use heuristics also suggests that interface-level constraints and usability rules can be used to improve agent reliability~\citep{liu2026augmenting}. Therefore, a promising next step is to build an execution layer that maps Brick Composer's symbolic outputs, such as brick identity and target pose, into verified GUI action sequences in a digital assembly environment.

\paragraph{Toward Grounded Robotic and VLA-Based Assembly.}
Finally, Brick Composer may serve as a step toward grounded physical assembly, but this direction should be pursued through careful intermediate stages. Robotic assembly requires not only part selection and pose estimation, but also procedure planning, grasp planning, contact reasoning, connector insertion, error recovery, and physical feedback. Prior work on language-first procedure planning shows that language can provide a useful intermediate structure for decomposing high-level goals into ordered executable steps~\citep{liu2023language}. Related work such as Manual2Skill further shows how visual instruction manuals can be converted into hierarchical assembly graphs and pose-conditioned execution plans for furniture assembly~\citep{tie2025manual2skill}, while Manual2Skill++ emphasizes connector-aware representations as a central requirement for general assembly~\citep{tie2025manual2skillpp}. More broadly, SayCan and RT-2 demonstrate that language and vision-language models can be connected to feasible actions through affordance grounding or vision-language-action training~\citep{ahn2022saycan,zitkovich2023rt2}. These works suggest a realistic path for extending Brick Composer: first use digital environments to verify part-level and pose-level decisions, then integrate language-based procedural decomposition, connector-aware representations, and physical feasibility checks, and only then transfer selected skills to robotic or VLA-based systems. In this way, future work can move from visual assembly understanding toward executable assembly while maintaining a clear grounding in physical constraints.

\end{document}